\documentclass[10pt,twocolumn,letterpaper]{article}

\usepackage[pagenumbers]{iccv} 

\definecolor{iccvblue}{rgb}{0.21,0.49,0.74}
\usepackage[pagebackref,breaklinks,colorlinks,allcolors=iccvblue]{hyperref}

\usepackage{caption}
\usepackage{subcaption}
\usepackage{bbding}
\usepackage{graphicx}
\usepackage{multirow}
\usepackage{amsmath} 
\usepackage{amssymb}
\usepackage{CJKutf8}
\usepackage[linesnumbered,ruled,vlined]{algorithm2e}
\newcommand{\halfcmark}{\cmark\kern-1.2ex\raisebox{0.7ex}{\rotatebox[origin=c]{125}{\textbf{--}}}}
\AtEndPreamble{
    \usepackage[capitalize]{cleveref}
    \crefname{section}{Sec.}{Secs.}
    \Crefname{section}{Section}{Sections}
    \Crefname{table}{Table}{Tables}
    \crefname{table}{Tab.}{Tabs.}
}
\definecolor{mygreen}{rgb}{0.0, 0.5, 0.0}  
\definecolor{myblue}{rgb}{0.0, 0.0, 0.7}   
\definecolor{myred}{rgb}{0.8, 0.0, 0.0}

\usepackage{colortbl}
\definecolor{mygray}{gray}{0.95}
\definecolor{my_green}{RGB}{82,208,80}
\usepackage{makecell}
\definecolor{00red}{RGB}{236,35,35}

\usepackage{pifont}
\newcommand{\cmark}{\ding{51}}%
\newcommand{\xmark}{\ding{55}}%

\def\paperID{****} 
\def\confName{ICCV}
\def\confYear{2025}

\title{MMReason: An Open-Ended Multi-Modal Multi-Step \\ Reasoning Benchmark for MLLMs Toward AGI}

\author{%
Huanjin Yao$^{1}$, ~
Jiaxing Huang $^{1}\textsuperscript{\Envelope}$, ~
Yawen Qiu$^{2}$, ~
Michael K. Chen$^{1}$, ~
Wenzheng Liu$^{4}$, ~
Wei Zhang$^{5}$ \\
Wenjie Zeng$^{2}$, ~
Xikun Zhang$^{1}$ , ~
Jingyi Zhang$^{1}$ , ~
YuXin Song$^{3}$, ~
Wenhao Wu$^{3}$, ~
Dacheng Tao$^{1}$\\
$^1$Nanyang Technological University \quad $^2$Tsinghua University \quad $^3$Baidu Inc.\\
$^4$ University of California \quad $^5$ University of Science and Technology of China \\
{\small \Envelope~Corresponding author} 
}

\begin{document}
\maketitle

\begin{abstract}
Reasoning plays a crucial role in advancing Multimodal Large Language Models (MLLMs) toward Artificial General Intelligence.
However, existing MLLM benchmarks often fall short in precisely and comprehensively evaluating long-chain reasoning abilities from three key aspects: (1) lack of difficulty and diversity, (2) susceptibility to guessability and memorization, (3) inadequate assessment of intermediate reasoning steps.
To fill this gap, we introduce \textbf{MMReason}, a new benchmark designed to precisely and comprehensively evaluate MLLM long-chain reasoning capability with diverse, open-ended, challenging questions.
First, we curate challenging questions requiring multi-step reasoning from various fields (i.e., 6 disciplines) and multiple difficulty levels (i.e., from pre-university to university, and from foundational to competition tiers).
Second, these questions are reformulated into an open-ended format and filtered using a multi-model voting technique to eliminate shortcut cases related to guessing and memorization, ensuring robust reasoning evaluations.
Third, we annotate the questions with detailed step-by-step solutions, and design a reference-based ternary scoring mechanism to reliably assess intermediate reasoning steps.
With MMReason, we benchmark popular leading MLLMs and provide an in-depth analysis of their reasoning capabilities.
We hope MMReason will serve as a valuable resource for advancing MLLM reasoning research.
Code will be available at \url{https://github.com/HJYao00/MMReason}.
\end{abstract}

\section{Introduction}
\label{sec:intro}

\begin{figure}[t]
    \centering
    \begin{subfigure}[b]{0.5\textwidth}
        \centering
        \includegraphics[width=\textwidth]{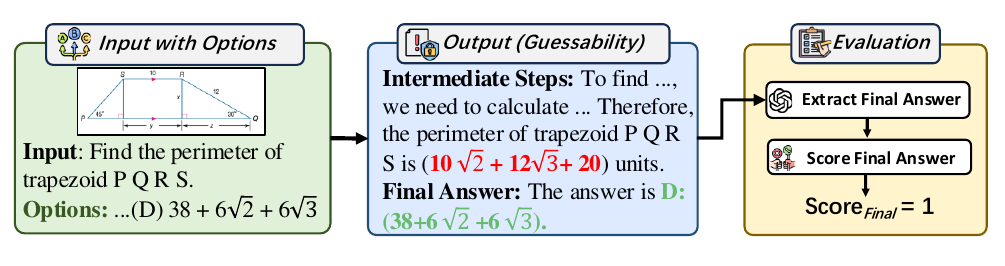}
        \caption{Guessability issues in existing benchmarks.}
        \label{fig:image1}
    \end{subfigure}
    \begin{subfigure}[b]{0.5\textwidth}
        \centering
        \includegraphics[width=\textwidth]{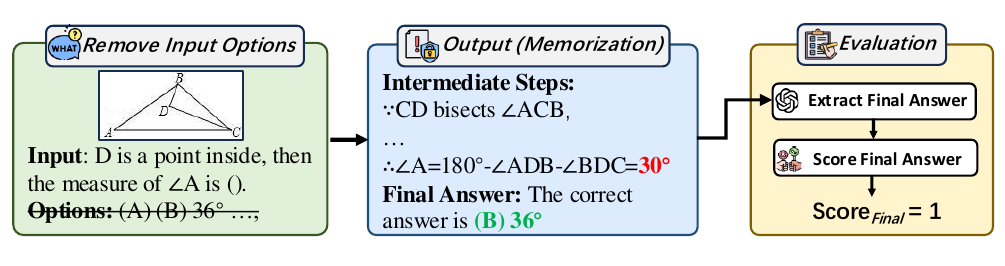}
        \caption{Memorization issues in existing benchmarks.}
        \label{fig:image2}
    \end{subfigure}

    \vspace{0cm}  

    \begin{subfigure}[b]{0.5\textwidth}
        \centering
        \includegraphics[width=\textwidth]{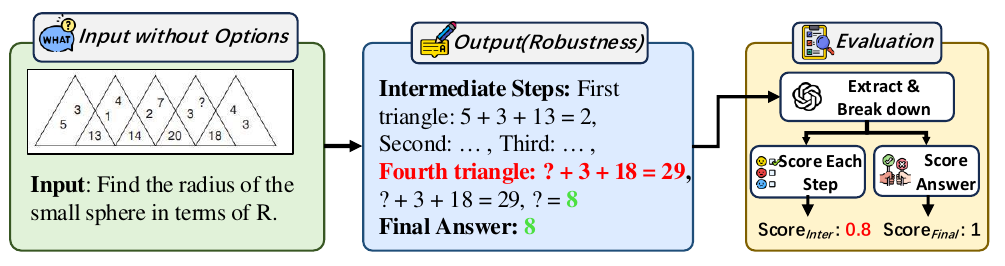}
        \caption{New robust questions and methods for long-chain reasoning evaluation.}
        \label{fig:image3}
    \end{subfigure}

    \caption{
    Existing benchmarks are susceptible to (a) guessability, where MLLMs arrive at correct answer by guessing despite flawed reasoning, and (b) memorization, where models recall leaked question types and answers rather than reasoning.
    We propose (c) a robust open-ended benchmark designed to evaluate long-chain reasoning by assessing both intermediate steps and final results.
    }
    \label{fig:motivation}
    \vspace{-0.5cm}
\end{figure}

Recent breakthroughs in the reasoning capabilities of Large Language Models (LLMs)~\cite{deepseek-r1, openai-o1, logic-rl, deepseek-v3, qwq} are progressively paving the way toward AGI. 
Inspired by these advances, Multimodal Large Language Models (MLLMs)~\cite{llava-cot, llava-reasoner, mulberry, qvq} have shifted their focus toward long-chain reasoning, and aim to integrate textual and visual information to generate long-chain reasoning steps, ultimately solving complex problems across various domains.
Leading-edge MLLMs, such as GPT-o1~\cite{openai-o1}, have exhibited exceptional understanding and reasoning abilities, even surpassing human performance. For example, on MathVista~\cite{mathvista}, GPT-o1 achieves a score of 73.9\%, which significantly outperforms the human performance of 60.3\%.
Despite these advancements, a critical question arises: \textit{how well do existing benchmarks truly evaluate the reasoning abilities of MLLMs?}

\begin{table*}[ht]
  \centering
  \scalebox{0.8}{
  \setlength{\tabcolsep}{4.5pt}
  \begin{tabular}{@{}lccccccccccc@{}}
    \toprule
    \multirow{2}{*}{Benchmark} & \multicolumn{4}{c}{Question Source (Grade \& Difficulty)}  & \multirow{2}{*}{\shortstack{Diverse\\Disciplines}} & \multirow{2}{*}{\shortstack{Questions \\ Filtering}} & \multirow{2}{*}{\shortstack{Intermediate Steps \\ Evaluation}} \\
    \cline{2-3} \cline{4-5} 
    & \multirow{1}{*}{pre-university} & \multirow{1}{*}{university} & Foundational & Competition &   &   & \\
    \midrule
    MathVista~\cite{mathvista} & \textcolor{mygreen}{\cmark} & \textcolor{myred}{\xmark} & \textcolor{myred}{\xmark}  & \textcolor{myred}{\xmark} & \textcolor{myred}{\xmark} & \textcolor{myred}{\xmark} & \textcolor{myred}{\xmark}\\
    MMMU~\cite{mmmu} & \textcolor{myred}{\xmark} & \textcolor{mygreen}{\cmark} & \textcolor{mygreen}{\cmark} & \textcolor{myred}{\xmark} & \textcolor{mygreen}{\cmark} & \textcolor{myred}{\xmark} & \textcolor{myred}{\xmark} \\
    MM-Math & \textcolor{mygreen}{\cmark} & \textcolor{myred}{\xmark} & \textcolor{mygreen}{\cmark} & \textcolor{myred}{\xmark} & \textcolor{myred}{\xmark} & \textcolor{myred}{\xmark} & \textcolor{myred}{\xmark} \\
    OlympiadBench~\cite{olympiadbench} & \textcolor{mygreen}{\cmark} & \textcolor{myred}{\xmark} & \textcolor{myred}{\xmark} & \textcolor{mygreen}{\cmark} & \textcolor{myred}{\xmark}  & \textcolor{myred}{\xmark} & \textcolor{myred}{\xmark} \\
    MME-CoT~\cite{mme-cot} & \textcolor{mygreen}{\cmark} & \textcolor{myred}{\xmark} & \textcolor{mygreen}{\cmark} & \textcolor{myred}{\xmark} & \textcolor{myred}{\xmark}  & \textcolor{myred}{\xmark} & \textcolor{mygreen}{\cmark} \\
    \midrule
    MMReason (Ours) & \textcolor{mygreen}{\cmark} & \textcolor{mygreen}{\cmark} & \textcolor{mygreen}{\cmark} & \textcolor{mygreen}{\cmark} & \textcolor{mygreen}{\cmark} & \textcolor{mygreen}{\cmark} & \textcolor{mygreen}{\cmark} \\ 
    \bottomrule
  \end{tabular}}
  \caption{Comparison of our MMReason with existing multimodal benchmarks}
  \label{tab: }
  \vspace{-1.5em}
\end{table*}

While existing benchmarks~\cite{mmvet,mmbench,mathvista,mme,mmmu,pope,hallusionbench} have laid the foundation for early-stage MLLM research and development, they often fall short in precisely and comprehensively evaluating MLLM long-chain reasoning capabilities. We identify three key limitations:
(1) \textit{Limited Difficulty and Diversity.}
Most existing MLLM benchmarks consist primarily of simple textual questions paired with easily interpretable images from limited domains, posing minimal challenges for today’s advanced models.
These homogeneous, low-complexity questions require little long-chain reasoning, making them insufficient for comprehensively assessing MLLMs' long-chain reasoning capabilities.
(2) \textit{Susceptibility to Guessability and Memorization.}
Most existing MLLM benchmarks are dominated by multiple-choice questions (MCQ), e.g., 94\% of MMMU is MCQ.
However, as illustrated in Figure~\ref{fig:motivation}, we observe that MCQ questions are susceptible to guessing and memorization issues.
Specifically, MLLMs may arrive at the correct final answer either via random guessing or by recalling leaked and memorized benchmark data, even when their intermediate reasoning steps are incorrect.
These shortcuts undermine the reliability of existing benchmarks in precisely measuring true long-chain reasoning capabilities.
(3) \textit{Lack of Intermediate Steps Evaluation.}
Most existing MLLM benchmarks assess only the final answer, neglecting the evaluation of intermediate reasoning steps, especially for multi-disciplinary exam questions that require complex, multi-step reasoning and critical thinking. This oversight prevents a thorough assessment of MLLMs' step-by-step reasoning processes.

To tackle these challenges, we propose MMReason, an open-ended, Multi-Modal, and Multi-Step Reasoning benchmark designed for precisely and comprehensively evaluating MLLM long-chain reasoning capabilities. MMReason is characterized by three key aspects:

\begin{enumerate}[label=\arabic*.]
    \item \textbf{\textit{Collect challenging and diverse reasoning questions.}}
     We first reuse existing benchmarks~\cite{m3cot, mmmu_pro, mmmu, mmstar} by carefully selecting and collecting questions that requires long-chain reasoning from them. 
     On the other hand, given the limited difficulty and diversity of these benchmarks, we further curate more challenging questions by sourcing new ones from the internet across various education grades and difficulty levels, covering pre-university school exams, Olympiad competitions, university exams, and university-level competitions.
     These questions span six widely studied disciplines (i.e., Mathematics, Business, Science, Engineering, Social Science, and Health), encompassing both foundational and competition-level topics. 
     Solving them requires critical thinking and multi-step reasoning, making them well-suited for assessing MLLMs' advanced long-chain reasoning capabilities.

\item \textbf{\textit{How can we mitigate imprecise reasoning evaluation caused by guessability and memorization?}} 
To reduce evaluation biases stemming from these issues, we employ two data-filtering strategies.
First, since the collected data includes multiple-choice questions, we select those with unambiguous answers and manually reformulate them into open-ended questions to minimize guessability.
Second, to address memorization issues, we design a multi-model voting technique to filter out leaked and memorized questions. This approach also enhances the multimodal relevance of the collected questions.
In this way, if a MLLM can still derive the correct final answer after potential shortcuts (e.g., random guessing from choices and recalling leaked answers) have been eliminated, we believe its intermediate reasoning steps are much more likely to be valid.
This, in turn, enables MMReason to provide a more precise assessment of MLLMs' final answer reasoning abilities while also indirectly reflecting their intermediate reasoning processes.

\item \textbf{\textit{How can we evaluate intermediate reasoning steps precisely and reliably?}} 
We annotate questions with detailed step-by-step solutions and design a reference-based ternary scoring mechanism to reliably evaluate MLLMs' intermediate reasoning steps.
Specifically, we use GPT-4o to break down model responses into key steps, and prompt GPT-4o to score each step with the aid of annotated references, categorizing them as correct, unverifiable, or incorrect.
With these references, this process mirrors a human grader evaluating a student's problem-solving steps.
Moreover, since some problems have multiple valid solutions, a grader may be unable to determine a step's correctness. In such cases, assigning a neutral score is a more reasonable choice.
Through this approach, MMReason provides a more precise and reliable evaluation of MLLMs' intermediate reasoning ability.

\end{enumerate}

With MMReason, we evaluate the long-chain reasoning abilities of popular and leading MLLMs from the perspectives of both intermediate reasoning steps and the reasoned final answer. Our contributions are summarized as follows:

(1) We curate new challenging and diverse questions across multiple grade levels and difficulty tiers to comprehensively assess MLLMs' long-chain reasoning abilities.

(2) We investigate the issues of guessability and memorization in existing multimodal benchmarks. To address them, we introduce two data-filtering strategies to eliminate shortcuts (\eg, random guessing from choices and recalling leaked answers), aiming to precisely evaluate the multi-step reasoning capabilities of MLLMs.

(3) We annotate questions with detailed step-by-step solutions and design a reference-based ternary scoring mechanism to reliably evaluate MLLMs’ intermediate steps.

(4) We evaluate various leading MLLMs with MMReason, demonstrating it is a challenging benchmark for long-chain reasoning. Notably, the best model GPT-4o~\cite{gpt4o} only achieves 25.7\% final answer reasoning accuracy and xx.x\% intermediate steps reasoning score.

\begin{figure*}[ht]
  \centering
   \includegraphics[width=1\linewidth]{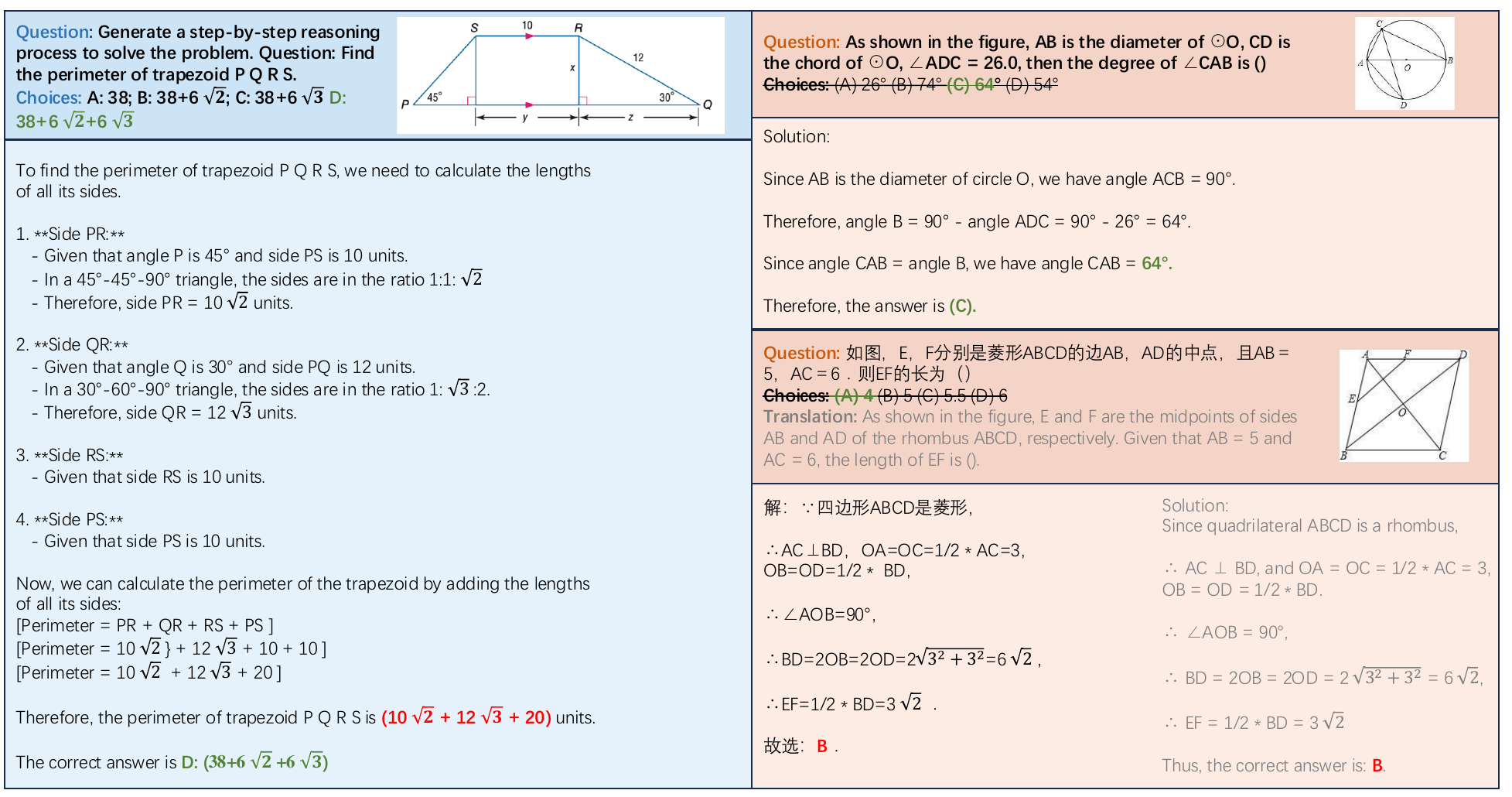}
   \caption{Examples of \textcolor[rgb]{0.1, 0.5, 0.8}{Guessability (Left)} and \textcolor[rgb]{0.68, 0.4, 0.2}{Memorization (Right)} issues. In the left figure, the model arrives at an incorrect numerical answer through multi-step reasoning but ultimately guesses the correct option by chance. In the right figure, we remove the choices from the original question before inputting it into MLLMs, yet the model still generates an answer choice, indicating the presence of memorization issues in existing benchmark.}
   \label{fig: fig2}
   \vspace{-0.5em}
\end{figure*}

\section{Related Work}

\subsection{Reasoning MLLMs}
The rapid growth of MLLMs~\cite{deepseek-vl2, Qwen2.5-VL, internvl2.5-mpo, minicpm-v, llava-ov, llama3, gpt4o, claude-3.5, gemini-1.5, zhang2024mm1} has endowed them with extensive knowledge and robust multitasking capabilities, enabling their application across complex and diverse domains.
More recently, inspired by advancements in reasoning within LLMs~\cite{openai-o1, deepseek-r1, qwq, team2025kimi}, \eg, Openai-o1's~\cite{openai-o1} deliberative thinking time before responding and DeepSeek R1's~\cite{deepseek-r1} large-scale reinforcement learning for enhanced reasoning, MLLMs have increasingly shifted their focus toward reasoning as well~\cite{system1tosys2, qvq, mulberry, llava-reasoner, llava-cot, r1v, r1-vl, R1-ShareVL, vision-r1, vlm-r1, xreasoner, wang2025visuothink}.
Several works~\cite{llava-cot, llava-reasoner, visualcot}, \eg, LLaVA-CoT~\cite{llava-cot}, leverage chain-of-thought to enhance MLLMs reasoning capabilities.
Others~\cite{astar, mulberry} employ MCTS-based methods to strengthen their reasoning abilities, such as Mulberry~\cite{mulberry} which introduces collective knowledge to MCTS to search reasoning paths, thereby enhancing reasoning and reflection capabilities.
Moreover, some open-source projects~\cite{r1v, wang-2025-open-r1-video} apply reinforcement learning to multimodal field, following DeepSeek-R1~\cite{deepseek-r1}.
The advancement of reasoning MLLMs prompts us to develop a more precise and comprehensive multimodal benchmark for evaluating their intermediate and final reasoning capabilities.

\subsection{Multimodal Benchmark}
Previous benchmarks~\cite{mme, mmbench, mme-cot, mmmu, mmstar, mmvet, pope, hallusionbench, mmmu_pro, dynamath, gqa, vqa, mathverse, masry2022chartqa, charxiv, mmtbench, olympiadbench, m3cot} effectively evaluate the performance of earlier MLLMs from various perspectives, including perception, cognition, and hallucination, laying the groundwork for their development.
However, as MLLMs rapidly advance in reasoning, existing benchmarks have inevitably become simplistic and homogeneous, with many dominated by multiple-choice and true/false questions, which limits their effectiveness in providing precise and comprehensive evaluations.
Recently, MMMU~\cite{mmmu} evaluates MLLMs using multi-disciplinary university exam questions but remains heavily reliant on multiple-choice questions.
OlympiadBench~\cite{olympiadbench} introduces a more challenging competition-level benchmark but is limited to mathematics and physics problems.
MMStar~\cite{mmstar} mitigates data leakage in existing benchmarks but also overlooks shortcuts in the questions.
More recently, MME-CoT~\cite{mme-cot} benchmarks chain-of-thought reasoning in MLLMs across three dimensions: quality, robustness, and efficiency.
In comparison, our proposed benchmark covers both foundational and competition-level problems, consisting of newly collected and reformulated questions, spanning various educational levels and 6 diverse disciplines.
Besides, we observe that guessability and memorization issues exist in most current benchmarks, affecting the precision of evaluations. To address them, we first conduct an in-depth study in Section~\ref{subsec: 3.1}, then propose optimization methods by removing shortcuts (\ie, options) in questions and applying multi-model voting-based filtering.
Additionally, we employ a reference-based ternary scoring mechanism to reliably assess intermediate reasoning steps.

\section{Analysis and Motivation}
\label{subsec: 3.1}
In this section, we delve into the issues of \textbf{guessability} and \textbf{memorization} in multimodal benchmarks, which hinder the effective evaluation of MLLM reasoning abilities.
Additionally, we provide illustrations and experimental results to support our observations.

(1) \textbf{Guessability.} Current multimodal benchmarks predominantly feature multiple-choice and true/false questions due to their ease of collection and evaluation.
However, questions containing shortcuts [(i.e., candidate true answers)] have inherent flaws that limit their ability to effectively evaluate the reasoning capabilities of MLLMs.
As illustrated in Figure~\ref{fig: fig2} (Left), during the reasoning process, the model generates incorrect reasoning but still selects the correct answer.
This suggests that MLLMs have a high likelihood of guessing the correct answer [among the several candidate answers] rather than reasoning, leading to imprecision in reasoning evaluation.

(2) \textbf{Memorization.} Different from previous studies~\cite{mmstar} that analyze unintentional data leakage in current benchmarks, this paper conducts a more in-depth study by removing shortcuts to further illustrate the issue of memorization in evaluation.
With the rapid development of MLLMs, the amount of data they are exposed to and their memorization ability continue to grow. Therefore, it is inevitable that models encounter similar questions and memorize their answers during training. 
As shown in Figure~\ref{fig: fig2} (right), we remove the options from the input, but MLLMs still generate the correct option, indicating that they have already memorized the answers rather than learning to reason, leading to the correct answer.

To support our observations, we conduct experiments on existing benchmarks by comparing the original accuracy with the accuracy after removing these shortcuts, as shown in Table~\ref{tab: remove mcq}. Experimental results show that when options in existing benchmarks are removed, MLLMs fail to produce the correct answers, even for questions they originally answered correctly. 
Notably, the performance of Qwen2-VL-7B on MathVista dropped sharply from 58.2\% to 37.8\%.
This indicates that MLLMs have not truly learned to reason, and in most cases, the intermediate responses generated by MLLMs are incorrect. 
Therefore, this motivates us to construct a more effective benchmark for precisely evaluating the multi-step reasoning capabilities of MLLMs.

\begin{table}[t]
  \centering
  \resizebox{0.99\linewidth}{!}{
  \setlength{\tabcolsep}{11.0pt}
  \begin{tabular}{@{}lc|ccc@{}}
    \toprule
    Benchmark & Input & GPT-4o & Qwen2-VL-7B \\
    \midrule
    \multirow{2}{*}{MathVista~\cite{mathvista}} & \{Q, C, I\} & 63.8 & 58.2  \\
     & \{Q, I\} & 40.8  & 37.8    \\
    \bottomrule
  \end{tabular}}
  \caption{
  Performance Comparison of Removing Choices from Existing Benchmarks. By removing answer choices, instances of guessability and memorization are reduced as MLLMs are required to reason and conclude the answers by themselves instead of simply picking one, leading to a more precise evaluation of MLLMs' long-chain reasoning capabilities.
  }
  \label{tab: remove mcq}
  \vspace{-1em}
\end{table}

\section{MMReason}
In this section, we provide a detailed introduction to the construction process and evaluation strategy of MMReason.
Section~\ref{subsec: 3.2} elaborates on our dataset curation process, while Section~\ref{subsec: 3.3} introduces our multi-model voting filtering method for selecting unmemorized and visually relevant data.
Finally, we outline the strategies for evaluating intermediate and final reasoning capability in Section~\ref{subsec: evaluation strategy}.

\subsection{Data Curation Process}
\label{subsec: 3.2}
To construct a new benchmark for precisely and comprehensively evaluating multi-step reasoning capability of MLLMs with diverse and challenging open-ended questions, we collect and reformulate questions from existing benchmarks and web, and annotate the questions with reference solutions. The details are as follows:

\noindent\textbf{Existing Benchmark Collection.}
Existing multimodal benchmarks include a vast collection of multiple-choice questions spanning various domains.
However, directly converting multiple-choice questions from existing benchmarks into open-ended formats could lead to inaccuracies in evaluation, particularly due to multiple valid answers and variations in wording.
To address this issue, we select several challenging benchmarks (\ie, MMMU, MMMU-Pro, MMStar, and M$^3$CoT) and identify questions with unique answers (\eg, numerical calculations, chromosomal compositions, and chemical compounds), while discarding questions like ``Which of the following statements is correct?''.
We then reformulate these questions into an open-ended format.

\noindent\textbf{New Data Collection.}
To further systematically collect new diverse and challenging open-ended questions,
we first review mainstream subjects across different grade levels, prioritizing those that require multi-step reasoning and extensive vision-text interaction. Based on this, we identify six disciplines, \ie, Math, Business, Science, Engineering, Social Sciences, and Health.
We then source both foundational and competition-level questions for each discipline from the Internet, spanning pre-university to university levels.
For foundational problems, we prioritize diversity, knowledge coverage, and critical thinking, while for competition-level questions, we focus on the necessity of visual information given their inherent difficulty.
Moreover, we intentionally gather a wide range of open-ended questions types (\eg, when, what, how) to enhance question diversity and assess MLLMs' reasoning ability across different question formats.
Finally, since most collected questions are in multiple-choice format and language-only, we extract multimodal questions and manually reformulate them into an open-ended format, ensuring each question has a unique answer.

\noindent\textbf{Intermediate Steps Annotation}
To reliably evaluate intermediate reasoning steps generated by MLLMs, we introduce a reference-based ternary scoring mechanism with detailed strategies outlined in Section~\ref{subsec: evaluation strategy}.
To implement the reference mechanism, we annotate questions with multi-step solutions. Specifically, we carefully select a subset of questions from our newly collected data across different disciplines and manually annotate the intermediate steps with GPT assistance to ensure accuracy at each step.

In total, we collect 2185 instances from existing benchmarks, which, combined with the 756 newly collected instances, results in a total of 2941 instances.

\subsection{Robustness Enhancement via Voting}
\label{subsec: 3.3}
To further mitigate memorization issues and enhance the visual relevance, we employ a multi-model voting mechanism with $K$ powerful MLLMs to filter out potentially memorized or visually weakly relevant instances from our collected multimodal questions $Q$, as detailed in algorithm~\ref{alg: voting}.
Specifically, we leverage a group of powerful MLLMs, denoted as $\{\pi_{1}, \pi_{2},...,\pi_{K}\}$, which perform filtering through voting iteratively for $T$ rounds.
During evaluation, we only input the open-ended textual questions while excluding the image to test whether the MLLMs can answer correctly without visual input.
If the model can still answer correctly without any shortcuts or visual cues, we consider the instance potentially memorized or visually weakly relevant.
After each round of evaluation by all MLLMs, we remove any correctly answered questions and proceed with the updated question set $Q'$ for the next round of filtering.
After $T$ iterations, we obtain 1384 robust and vision-related multimodal questions, forming MMReason.

\begin{algorithm}[t]
\caption{Multi-modal Voting filtering}
\label{alg: voting}
\KwIn{a group of MLLMs $\{\pi_{1}, \pi_{2},...,\pi_{K}\}$; A set of multimodal questions $Q = \{T, V\}$ containing $M$ questions.}

  \For{i = 1 to T}{
    \For{k = 1 to K}{
        \textcolor{gray}{Input only text $T$ during evaluation:} \\
        Evaluation($\pi_k$, $Q_T=\{T\}$) \\
    }
    \textcolor{gray}{Traverse the question set Q:} \\
    \For{j = 1 to M}{
        \textcolor{gray}{If question $q_j$ is answered correctly more than 0 times , remove it:} \\
        \If{Count($q_j$) $>$ 0}{
            Remove $q_j$ from $Q$ \\
        }
    }
  }
  \KwOut{Filtered Data {$Q$}}
\end{algorithm}

\begin{table*}[ht]
  \centering\arraybackslash
  \scalebox{0.85}{
  \setlength{\tabcolsep}{8pt}
  \begin{tabular}{@{}lc|cc|cccccccccc@{}}
    \toprule
    \makecell[c]{\multirow{2.5}{*}{Models}} & \makecell[c]{\multirow{2.5}{*}{\shortstack{Size}  }} & \multicolumn{2}{c|}{Overall} & \multicolumn{6}{c}{Discipline-wise final reasoning accuracy}  \\
    \cmidrule(lr){3-4} \cmidrule(lr){5-10}
     &  & \multicolumn{1}{c}{Final} & Intermediate & Math & Business & Science  & Engineering & Social & Health  \\
    \midrule
    \multicolumn{9}{c}{ \em Closed-Source Models} \\ 
    \midrule
    GPT-4o-1120~\cite{gpt4o} & - & \textbf{25.7} & \textbf{42.1} & 33.3 & \textbf{36.8} & 14.5 & 8.8 & \textbf{42.1} & \textbf{47.6}  \\
    Claude-3.7V Sonnet~\cite{claude-3.5} & - & 25.1 & 36.1 & 40.0 & 30.4 & \textbf{18.6} & 10.0 & 22.4 & 34.9  \\
    Gemini-1.5 Pro~\cite{gemini-1.5} & - & 24.9 & 33.3 & \textbf{43.1} & 27.2 & 17.6 & \textbf{10.8} & 35.5 & 25.4 \\
    \midrule
    \multicolumn{9}{c}{\em Open-Source Models} \\
    \midrule
    Deepseek-VL2~\cite{deepseek-vl2} & 4.1B$^{\dag}$ & 12.9 &12.7 & 23.3 & 14.1 & 7.7 & 2.5 & 29.9 & 22.2 \\
    LLaVA-OneVision~\cite{llava-ov} & 7B & 9.6 & 7.9 & 24.3 & 8.5 & 3.8 & 1.2 & 2.6 & 3.2 \\
    Qwen-2.5-VL~\cite{internvl2.5-mpo} & 7B & 16.8 & 17.3 & 25.1 & 17.4 & 14.5 & 4.4 & 25.2 & 12.7 \\
    MiniCPM-V-2.6~\cite{minicpm-v} & 8B & 11.7 & 12.5 & 24.3 & 11.6 & 8.5 & 1.0 & 25.2 & 7.9 \\
    InternVL-2.5-MPO~\cite{internvl2.5-mpo} & 8B & 12.4 & 13.1 & 27.4 & 12.9 & 7.0 & 3.2 & 20.6 & 6.3  \\
    LLaMA-3.2-Vision~\cite{llama3} & 11B & 11.9 & 10.8 & 25.5 & 10.3 & 9.1 & 1.7 & 24.3 & 11.1 \\
    LLaVA-CoT~\cite{llava-cot} & 11B & 12.5 & 11.4 & 25.7 & 14.9 & 5.2 & 3.4 & 22.4 & 11.1 \\
    Mulberry~\cite{mulberry} & 11B & 12.9 & 12.3 & 26.7 & 13.6 & 7.7 & 2.0 & 27.1 & 9.6   \\
    InternVL-2.5-MPO~\cite{internvl2.5-mpo} & 78B & 21.3 & 23.8 & 33.1 & 23.7 & 17.6 & 6.6 & 33.6 & 28.6 \\
    Qwen-2.5-VL~\cite{Qwen2.5-VL} & 72B & 24.7 & 28.1 & 31.0 & 34.2 & \textbf{18.6} & 9.6 & 37.4 & 31.7 \\
    \bottomrule
  \end{tabular}}
  \caption{\textbf{Main Results.} We provide the final-answer reasoning accuracy and the intermediate-step reasoning scores for 13 mainstream MLLMs, as well as the final answer reasoning accuracy across different disciplines. $^{\dag}$ denotes activated parameters of MoE model.
  }
  \label{tab:Main Results}
  \vspace{-0.15em}
\end{table*}

\subsection{Evaluation Strategy}
\label{subsec: evaluation strategy}

\noindent\textbf{Intermediate step reasoning ability evaluation.}
A straightforward approach to evaluate intermediate steps is to utilize a highly capable MLLM (\eg, GPT-4o) to extract key steps from reasoning responses and directly assign binary scores to them.
However, directly using powerful MLLMs to score these segmented steps is unreliable, particularly for questions beyond model’s own knowledge.
To address this, we adopt a reference-based ternary scoring mechanism (\ie, correct, unverifiable, and incorrect), corresponding to scores of 0, 0.5, and 1, respectively, to ensure a more reliable assessment of intermediate reasoning steps.
Here, 'Unverifiable' refers to steps where MLLMs are still unable to make an accurate judgment despite having access to the reference solutions.
Specifically, we use GPT-4o to break down intermediate reasoning responses into $N$ steps, denoted as $[s_1, s_2, ..., s_N]$.
Next, we provide GPT-4o with referenced multi-step solutions and evaluate each step using a ternary score.
After scoring, we compute the final score by summing and averaging the scores across all steps:

\begin{equation}
    S_{inter} = \frac{1}{N}\sum_{n}^{N} Score(s_n)
\end{equation}

\noindent \textbf{Final answer reasoning ability evaluation.}
To assess final answer reasoning ability of MLLMs, we depart from the common approach of using a prompt like ``Answer the question using a single word or phrase.'' which forces MLLMs to output only the final result.
Instead, we encourage MLLMs to generate multi-step responses leading to the final answer, allowing us to evaluate their reasoning capabilities rather than memorizing. 
Since MMReason involves long-chain reasoning and unrestricted answer formats for each question, it is challenging to determine correctness through predefined rules.
Therefore, following previous methods~\cite{mathvista, mmmu, mathverse}, we prompt GPT-4o to extract the final answer from response and then assess its correctness, obtaining the final answer reasoning accuracy $S_{final}$.

\section{Experiments}

\begin{table}[t]
    \centering
    \scalebox{0.68}{
    \setlength{\tabcolsep}{2.0pt}
    \begin{tabular}{@{}lcccccc@{}}
    \toprule
    \multirow{2.5}{*}{Models}& \multicolumn{4}{c}{Reformulated from} & \multirow{2.5}{*}{\shortstack{Newly collect \\ (Ours)}} \\
    \cmidrule(lr){2-5}
    & M$^3$CoT & MMStar & MMMU\_Pro & MMMU &  \\
    \midrule
    DeepSeek-vl2~\cite{deepseek-vl2} & 16.9 & 22.7 & 11.6 & 12.4 & \textbf{8.3} \\
    InternVL2.5-78B-MPO~\cite{internvl2.5-mpo}& 29.1 & 28.6 & 20.1 & 18.6 & \textbf{14.2} \\
    Qwen2.5-VL-72B~\cite{Qwen2.5-VL} & 32.3 & 30.3 & 21.1 & 21.6 & \textbf{18.5}   \\ 
    GPT-4o-1120~\cite{gpt4o} & 28.9 & 32.7 & 24.9 & 24.5 & \textbf{21.8} \\
    \bottomrule
  \end{tabular}
  }
  \caption{\textbf{Final answer reasoning accuracy over different data sources.} We analyze the accuracy of different question sources in MMReason. Our newly collected data has the lowest accuracy, highlighting its difficulty for MLLMs.
  }
  \label{tab:score of different source}
  \vspace{-0.15em}
\end{table}

\begin{table}[t]
    \centering
    \scalebox{0.7}{
    \setlength{\tabcolsep}{3.0pt}
    \begin{tabular}{@{}lcccc@{}}
    \toprule
    Models & Text-only & Text \& Visual & Multimodal Relevance Rate\textbf{$\uparrow$}\\
    \midrule
    \textit{Before filtering} \\
    LLaMA-3.2-11B & 6.51 & 15.0 & 56.6\\
    InternVL2.5-8B-MPO& 7.23 & 17.9 & 59.6  \\
    Qwen2.5-VL-7B & 9.36 & 23.8 & 60.7 \\
    GPT4o & 13.4 & 30.8 & 56.5 \\
    \midrule
    \textit{After filtering}  \\
    LLaMA-3.2-Vision-11B & 0.65 & 11.7 & 94.5 \\
    InternVL2.5-8B-MPO & 0.69 & 12.4 & 94.5 \\
    Qwen2.5-VL-7B & 0.73 & 18.3 & 96.1\\
    GPT4o & 0.78 & 25.7 & 97.0 \\
    \bottomrule
  \end{tabular}
  }
  \caption{\textbf{Comparison before and after multi-model voting filtering.}  'Text' and 'Visual' indicate whether textual questions and visual information are provided, respectively. 
  }
  \label{tab:filter}
  \vspace{-0.15em}
\end{table}

\begin{figure*}[t]
  \centering
   \includegraphics[width=0.8\linewidth]{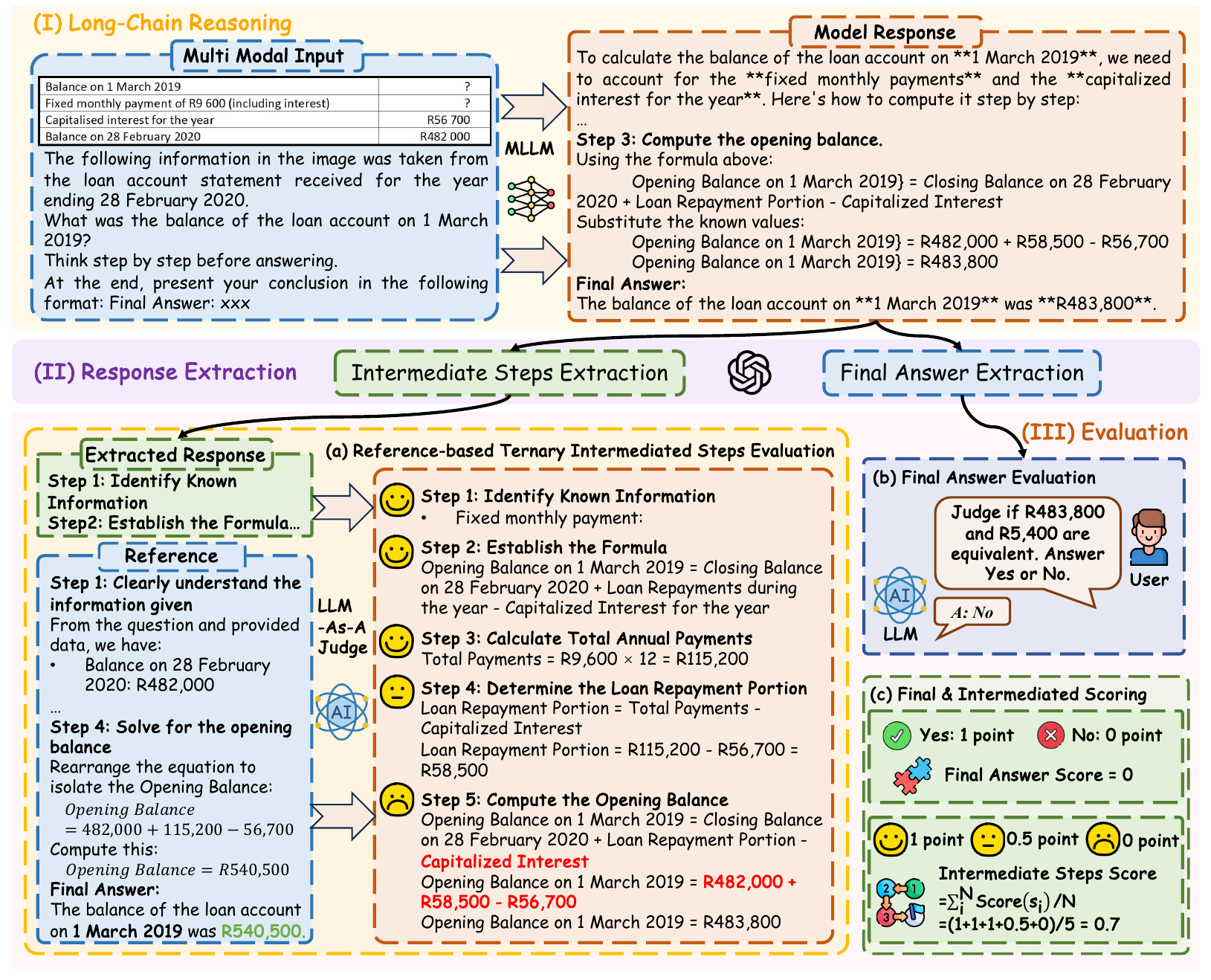}
   \caption{Qualitative analysis.}
   \label{fig: fig3}
   \vspace{-0.1em}
\end{figure*}
\vspace{-1em}

\subsection{Experiment Setup}

\textbf{Filtering Setting}
We implement $K=4$ well-known MLLMs for multi-model voting filtering to mitigate memorization issues and enhance visual relevance. These models include GPT-4o~\cite{gpt4o}, LLaMA3.2-Vision-11B~\cite{llama3}, InternVL2.5-8B-MPO~\cite{internvl2.5-mpo}, and Qwen2.5-VL-7B~\cite{Qwen2.5-VL}.
These MLLMs are widely studied by both industry and academic, including in data distillation and post-training. 
Therefore, leveraging them for filtering in $T=2$ rounds enhance the robustness for our benchmark.

\noindent \textbf{Evaluation Setting.}
We benchmark the reasoning capabilities of 13 leading MLLMs on MMReason, including 3 closed-source and 10 open-source models.
For closed-source MLLMs, we benchmark GPT-4o-1120~\cite{gpt4o}, Claude-3.7V Sonnet~\cite{claude-3.5} and Gemini-1.5 Pro~\cite{gemini-1.5}. For open-source models, we evaluate Deepseek-VL2~\cite{qvq}, LLaVA-OneVision-7B~\cite{llava-ov}, Mini-CPM-V-2.6-8B~\cite{minicpm-v}, LLaMA-3.2-Vision-90B-Instruct~\cite{llama3}, Mulberry~\cite{mulberry}, InterVL-2.5-MPO-8B\&78B~\cite{internvl2.5-mpo}, Qwen2.5-VL-7B\&72B\cite{Qwen2.5-VL}.
We use VLMEvalKit~\cite{vlmevalkit} to evaluate these models on MMReason.

\subsection{Experimental Results}

\subsubsection{Main Results}
In this section, we comprehensively compare the reasoning capabilities of various MLLMs in both intermediate steps and final answers on \textbf{MMReason}. Table~\ref{tab:Main Results} presents their overall final-answer reasoning accuracy and intermediate-step reasoning accuracy, with final-answer reasoning accuracy across different disciplines listed in right.

\noindent\textbf{Final answer reasoning accuracy.} MMReason benchmark presents a significant challenge to current MLLMs, requiring them to derive correct answers through multi-step reasoning without shortcuts (\ie, answer choices).
~Table~\ref{tab:Main Results} shows that even the highest accuracy, obtained by GPT-4o~\cite{gpt4o}, reaches only 25.7\%.
Notably, Qwen-2.5-VL-72B~\cite{Qwen2.5-VL} significantly outperforms other open-source models, achieving an accuracy of 24.7\%, which is comparable to proprietary models, \ie, 24.9\% on Gemini-1.5 Pro~\cite{gemini-1.5}, 25.1\% on Calude-3.7V Soonet~\cite{claude-3.5}.
For smaller reasoning models, the o1-like Mulberry-11B~\cite{mulberry} and the MoE-based DeepSeek-VL2 achieved the same accuracy of 12.9\%.
The accuracy of final answer reasoning reflects the challenging nature of our benchmark.

\noindent\textbf{Intermediate steps reasoning scores.}
We design a reference-based ternary scoring mechanism to evaluate the intermediate steps generated by the MLLMs, with the results shown in Table~\ref{tab:Main Results}. Among them, GPT-4o~\cite{gpt4o} achieves the highest score in intermediate step reasoning evaluation, with a score of 42.1\%. Compared to the accuracy of the final answer reasoning, closed-source models perform relatively better in intermediate step reasoning, showing a larger gap from open-source models. This demonstrates their stronger capability in long-chain reasoning.

\noindent\textbf{Discipline-wise final reasoning accuracy.} We investigate the accuracy of final answer reasoning across different disciplines.
The results show that Gemini-1.5 Pro exhibits the strongest mathematical reasoning ability, achieving 43.1\%.
Moreover, engineering-related questions are generally more challenging to answer. For example, MiniCPM-V-2.6 achieves only 1\% accuracy on these questions, as they require extensive domain knowledge and are most likely derived from university-level.
The diversity of questions in MMReason enables us to evaluate the reasoning abilities of different models across various disciplines.

\subsection{Analysis}
\textbf{Final answer reasoning accuracy from different sources.} In Table~\ref{tab:score of different source}, we provide accuracy across different data sources, including questions collected and reformulated from existing benchmarks, as well as newly collected ones.
The results are evaluated on three powerful reasoning models (\ie, LLaMA-3.2-vision-11B~\cite{llama3}, InternVL2.5-8B-MPO~\cite{internvl2.5-mpo}, Qwen2.5-VL-7B~\cite{Qwen2.5-VL}, and GPT-4o~\cite{gpt4o}), showing that our newly collected diverse data is the most challenging. This is attributed to our data collection strategy, which incorporates varying difficulty levels and educational stages across multiple disciplines.

\begin{figure}[t]
  \centering
   \includegraphics[width=0.9\linewidth]{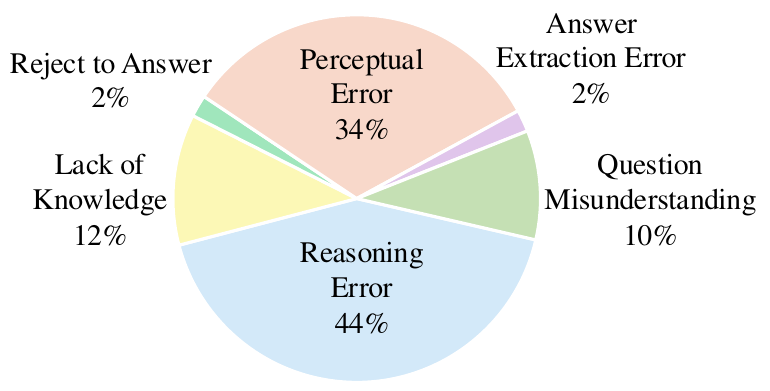}
   \caption{Error distribution over 50 incorrect responses from GPT-4o, with reasoning and perception errors being the most prevalent.}
   \label{fig: error distributions}
   \vspace{-0.5em}
\end{figure}

\noindent\textbf{Memorization.}
In Table~\ref{tab:filter}, we present the accuracy before and after filtering, with and without visual input. We also use the multi-modal relevance rate to analyze the extent of memorization in different models. 
Specifically, we first evaluate the 2941 initially open-ended questions under two conditions: text-only input and combined text-visual input.
The results in Table~\ref{tab:filter} indicate that before filtering, the collected dataset exhibits significant memorization effects or visual irrelevance, \ie, many of these questions can be answered correctly without requiring images. For example, GPT-4o achieves an accuracy of 13.4\% with text-only input.
Therefore, we remove these memorized data to enhance the robustness of our benchmark. After $T$ rounds of iterative filtering, we conduct another evaluation to ensure a higher multimodal relevance rate in MMReason.
These results indicate that the retained questions in MMReason are largely free from memorization and require visual input to be solved, ensuring that our benchmark precisely evaluates the long-chain reasoning capabilities of MLLMs.

\subsubsection{Qualitative analysis}

\textbf{Reasoning Analysis.}
Figure 3 represents the overall evaluation process of MMReason's Qualitative Results, including both final answer reasoning evaluation and intermediate step evaluation. 
For the final answer evaluation, we remove the shortcut (\ie, the options), preventing the model from guessing the answer and forcing it to derive the final answer step by step through reasoning. The results in the figure indicate that the model misunderstands ``Capitalized interests'', leading to incorrect reasoning. This demonstrates that MMReason can precisely evaluate the model's reasoning ability.
For intermediate step evaluation, the presence of references allows for more accurate assessment of the intermediate reasoning steps. Additionally, the use of a ternary scoring system provides a more reliable intermediate score for steps that the scoring model cannot confidently judge.

\noindent\textbf{Error Analysis.}
In this section, we analyze the errors made by powerful MLLM GPT-4o in its multi-step reasoning responses.
We hope this study will provide valuable insights to enhance the reasoning performance of MLLMs in the future.
Specifically, we randomly select 50 cases of final reasoning errors and analyzed their causes. We identified six potential factors contributing to these errors, \ie, Lack of Knowledge, Reject to Answer, Answer Extraction Error, Question Misunderstanding, Reasoning Error, and Perceptual Error, 
The error distribution is shown in Figure~\ref{fig: error distributions}, with additional examples provided in the Appendix.

\begin{figure}[t]
  \centering
   \includegraphics[width=1\linewidth]{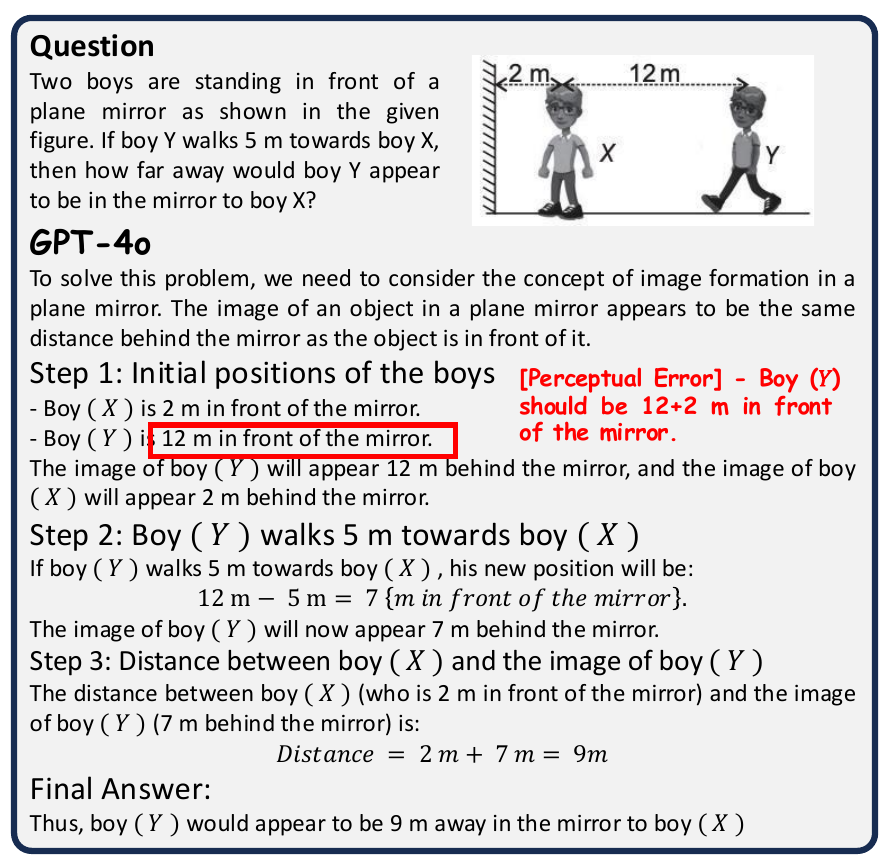}
   \caption{\textbf{Qualitative results of incorrect reasoning.}}
   \label{fig: error qualitative results}
   \vspace{-0.5em}
\end{figure}

\section{Conclusion}
In this paper, we introduce MMReason, a new benchmark aiming to precisely and comprehensively evaluate long-chain reasoning capability of MLLMs with diverse and challenging open-ended questions that require multi-step-reasoning and span various education grades and difficulty levels.
We evaluate the reasoning performance of multiple mainstream MLLMs on MMReason.
Notably, GPT-4o achieves the highest performance, yet its accuracy remains only 25.7\%. Besides, our analysis of GPT-4o's incorrect responses reveal that perception and reasoning error constitute a significant portion. 
These findings suggest that MLLMs still have considerable potential for improvement in reasoning.
We hope MMReason offers valuable insights to advance the reasoning capabilities of MLLMs.

\clearpage

{
    \small
    \bibliographystyle{ieeenat_fullname}
    \bibliography{main}

\begin{thebibliography}{51}
\providecommand{\natexlab}[1]{#1}
\providecommand{\url}[1]{\texttt{#1}}
\expandafter\ifx\csname urlstyle\endcsname\relax
  \providecommand{\doi}[1]{doi: #1}\else
  \providecommand{\doi}{doi: \begingroup \urlstyle{rm}\Url}\fi

\bibitem[Antol et~al.(2015)Antol, Agrawal, Lu, Mitchell, Batra, Zitnick, and Parikh]{vqa}
Stanislaw Antol, Aishwarya Agrawal, Jiasen Lu, Margaret Mitchell, Dhruv Batra, C~Lawrence Zitnick, and Devi Parikh.
\newblock Vqa: Visual question answering.
\newblock In \emph{Proceedings of the IEEE international conference on computer vision}, pages 2425--2433, 2015.

\bibitem[Bai et~al.(2025)Bai, Chen, Liu, Wang, Ge, Song, Dang, Wang, Wang, Tang, Zhong, Zhu, Yang, Li, Wan, Wang, Ding, Fu, Xu, Ye, Zhang, Xie, Cheng, Zhang, Yang, Xu, and Lin]{Qwen2.5-VL}
Shuai Bai, Keqin Chen, Xuejing Liu, Jialin Wang, Wenbin Ge, Sibo Song, Kai Dang, Peng Wang, Shijie Wang, Jun Tang, Humen Zhong, Yuanzhi Zhu, Mingkun Yang, Zhaohai Li, Jianqiang Wan, Pengfei Wang, Wei Ding, Zheren Fu, Yiheng Xu, Jiabo Ye, Xi Zhang, Tianbao Xie, Zesen Cheng, Hang Zhang, Zhibo Yang, Haiyang Xu, and Junyang Lin.
\newblock Qwen2.5-vl technical report.
\newblock \emph{arXiv preprint arXiv:2502.13923}, 2025.

\bibitem[Chen et~al.(2024{\natexlab{a}})Chen, Li, Dong, Zhang, Zang, Chen, Duan, Wang, Qiao, Lin, et~al.]{mmstar}
Lin Chen, Jinsong Li, Xiaoyi Dong, Pan Zhang, Yuhang Zang, Zehui Chen, Haodong Duan, Jiaqi Wang, Yu Qiao, Dahua Lin, et~al.
\newblock Are we on the right way for evaluating large vision-language models?
\newblock \emph{arXiv preprint arXiv:2403.20330}, 2024{\natexlab{a}}.

\bibitem[Chen et~al.(2025)Chen, Li, Zhao, Song, and Vinci]{r1v}
Liang Chen, Lei Li, Haozhe Zhao, Yifan Song, and Vinci.
\newblock R1-v: Reinforcing super generalization ability in vision-language models with less than \$3.
\newblock \url{https://github.com/Deep-Agent/R1-V}, 2025.
\newblock Accessed: 2025-02-02.

\bibitem[Chen et~al.(2024{\natexlab{b}})Chen, Qin, Zhang, Chen, Xu, and Che]{m3cot}
Qiguang Chen, Libo Qin, Jin Zhang, Zhi Chen, Xiao Xu, and Wanxiang Che.
\newblock M$^3$cot: A novel benchmark for multi-domain multi-step multi-modal chain-of-thought.
\newblock In \emph{Proc. of ACL}, 2024{\natexlab{b}}.

\bibitem[Duan et~al.(2024)Duan, Yang, Qiao, Fang, Chen, Liu, Dong, Zang, Zhang, Wang, et~al.]{vlmevalkit}
Haodong Duan, Junming Yang, Yuxuan Qiao, Xinyu Fang, Lin Chen, Yuan Liu, Xiaoyi Dong, Yuhang Zang, Pan Zhang, Jiaqi Wang, et~al.
\newblock Vlmevalkit: An open-source toolkit for evaluating large multi-modality models.
\newblock In \emph{Proceedings of the 32nd ACM international conference on multimedia}, pages 11198--11201, 2024.

\bibitem[Dubey et~al.(2024)Dubey, Jauhri, Pandey, Kadian, Al-Dahle, Letman, Mathur, Schelten, Yang, Fan, et~al.]{llama3}
Abhimanyu Dubey, Abhinav Jauhri, Abhinav Pandey, Abhishek Kadian, Ahmad Al-Dahle, Aiesha Letman, Akhil Mathur, Alan Schelten, Amy Yang, Angela Fan, et~al.
\newblock The llama 3 herd of models.
\newblock \emph{arXiv preprint arXiv:2407.21783}, 2024.

\bibitem[Fu et~al.(2024)Fu, Chen, Shen, Qin, Zhang, Lin, Yang, Zheng, Li, Sun, Wu, and Ji]{mme}
Chaoyou Fu, Peixian Chen, Yunhang Shen, Yulei Qin, Mengdan Zhang, Xu Lin, Jinrui Yang, Xiawu Zheng, Ke Li, Xing Sun, Yunsheng Wu, and Rongrong Ji.
\newblock Mme: A comprehensive evaluation benchmark for multimodal large language models, 2024.

\bibitem[Guan et~al.(2024)Guan, Liu, Wu, Xian, Li, Liu, Wang, Chen, Huang, Yacoob, et~al.]{hallusionbench}
Tianrui Guan, Fuxiao Liu, Xiyang Wu, Ruiqi Xian, Zongxia Li, Xiaoyu Liu, Xijun Wang, Lichang Chen, Furong Huang, Yaser Yacoob, et~al.
\newblock Hallusionbench: an advanced diagnostic suite for entangled language hallucination and visual illusion in large vision-language models.
\newblock In \emph{Proceedings of the IEEE/CVF Conference on Computer Vision and Pattern Recognition}, pages 14375--14385, 2024.

\bibitem[Guo et~al.(2025)Guo, Yang, Zhang, Song, Zhang, Xu, Zhu, Ma, Wang, Bi, et~al.]{deepseek-r1}
Daya Guo, Dejian Yang, Haowei Zhang, Junxiao Song, Ruoyu Zhang, Runxin Xu, Qihao Zhu, Shirong Ma, Peiyi Wang, Xiao Bi, et~al.
\newblock Deepseek-r1: Incentivizing reasoning capability in llms via reinforcement learning.
\newblock \emph{arXiv preprint arXiv:2501.12948}, 2025.

\bibitem[He et~al.(2024)He, Luo, Bai, Hu, Thai, Shen, Hu, Han, Huang, Zhang, et~al.]{olympiadbench}
Chaoqun He, Renjie Luo, Yuzhuo Bai, Shengding Hu, Zhen~Leng Thai, Junhao Shen, Jinyi Hu, Xu Han, Yujie Huang, Yuxiang Zhang, et~al.
\newblock Olympiadbench: A challenging benchmark for promoting agi with olympiad-level bilingual multimodal scientific problems.
\newblock \emph{arXiv preprint arXiv:2402.14008}, 2024.

\bibitem[Huang et~al.(2025)Huang, Jia, Zhai, Cao, Ye, Zhao, Xu, Hu, and Lin]{vision-r1}
Wenxuan Huang, Bohan Jia, Zijie Zhai, Shaosheng Cao, Zheyu Ye, Fei Zhao, Zhe Xu, Yao Hu, and Shaohui Lin.
\newblock Vision-r1: Incentivizing reasoning capability in multimodal large language models.
\newblock \emph{arXiv preprint arXiv:2503.06749}, 2025.

\bibitem[Hudson and Manning(2019)]{gqa}
Drew~A Hudson and Christopher~D Manning.
\newblock Gqa: A new dataset for real-world visual reasoning and compositional question answering.
\newblock In \emph{Proceedings of the IEEE/CVF conference on computer vision and pattern recognition}, pages 6700--6709, 2019.

\bibitem[Hurst et~al.(2024)Hurst, Lerer, Goucher, Perelman, Ramesh, Clark, Ostrow, Welihinda, Hayes, Radford, et~al.]{gpt4o}
Aaron Hurst, Adam Lerer, Adam~P Goucher, Adam Perelman, Aditya Ramesh, Aidan Clark, AJ Ostrow, Akila Welihinda, Alan Hayes, Alec Radford, et~al.
\newblock Gpt-4o system card.
\newblock \emph{arXiv preprint arXiv:2410.21276}, 2024.

\bibitem[Jaech et~al.(2024)Jaech, Kalai, Lerer, Richardson, El-Kishky, Low, Helyar, Madry, Beutel, Carney, et~al.]{openai-o1}
Aaron Jaech, Adam Kalai, Adam Lerer, Adam Richardson, Ahmed El-Kishky, Aiden Low, Alec Helyar, Aleksander Madry, Alex Beutel, Alex Carney, et~al.
\newblock Openai o1 system card.
\newblock \emph{arXiv preprint arXiv:2412.16720}, 2024.

\bibitem[Jiang et~al.(2025)Jiang, Zhang, Guo, Li, Qi, Chen, Wang, Jin, Guo, Yan, et~al.]{mme-cot}
Dongzhi Jiang, Renrui Zhang, Ziyu Guo, Yanwei Li, Yu Qi, Xinyan Chen, Liuhui Wang, Jianhan Jin, Claire Guo, Shen Yan, et~al.
\newblock Mme-cot: Benchmarking chain-of-thought in large multimodal models for reasoning quality, robustness, and efficiency.
\newblock \emph{arXiv preprint arXiv:2502.09621}, 2025.

\bibitem[Li et~al.(2024)Li, Zhang, Guo, Zhang, Li, Zhang, Zhang, Zhang, Li, Liu, et~al.]{llava-ov}
Bo Li, Yuanhan Zhang, Dong Guo, Renrui Zhang, Feng Li, Hao Zhang, Kaichen Zhang, Peiyuan Zhang, Yanwei Li, Ziwei Liu, et~al.
\newblock Llava-onevision: Easy visual task transfer.
\newblock \emph{arXiv preprint arXiv:2408.03326}, 2024.

\bibitem[Li et~al.(2023)Li, Du, Zhou, Wang, Zhao, and Wen]{pope}
Yifan Li, Yifan Du, Kun Zhou, Jinpeng Wang, Wayne~Xin Zhao, and Ji-Rong Wen.
\newblock Evaluating object hallucination in large vision-language models.
\newblock \emph{arXiv preprint arXiv:2305.10355}, 2023.

\bibitem[Li et~al.(2025)Li, Zhang, Zhang, Zhang, Liu, Yao, Xu, Zheng, Wang, Chen, et~al.]{system1tosys2}
Zhong-Zhi Li, Duzhen Zhang, Ming-Liang Zhang, Jiaxin Zhang, Zengyan Liu, Yuxuan Yao, Haotian Xu, Junhao Zheng, Pei-Jie Wang, Xiuyi Chen, et~al.
\newblock From system 1 to system 2: A survey of reasoning large language models.
\newblock \emph{arXiv preprint arXiv:2502.17419}, 2025.

\bibitem[Liu et~al.(2024{\natexlab{a}})Liu, Feng, Xue, Wang, Wu, Lu, Zhao, Deng, Zhang, Ruan, et~al.]{deepseek-v3}
Aixin Liu, Bei Feng, Bing Xue, Bingxuan Wang, Bochao Wu, Chengda Lu, Chenggang Zhao, Chengqi Deng, Chenyu Zhang, Chong Ruan, et~al.
\newblock Deepseek-v3 technical report.
\newblock \emph{arXiv preprint arXiv:2412.19437}, 2024{\natexlab{a}}.

\bibitem[Liu et~al.(2025)Liu, Zhang, Qin, Ossowski, Gu, Jin, Kiblawi, Preston, Wei, Vozila, et~al.]{xreasoner}
Qianchu Liu, Sheng Zhang, Guanghui Qin, Timothy Ossowski, Yu Gu, Ying Jin, Sid Kiblawi, Sam Preston, Mu Wei, Paul Vozila, et~al.
\newblock X-reasoner: Towards generalizable reasoning across modalities and domains.
\newblock \emph{arXiv preprint arXiv:2505.03981}, 2025.

\bibitem[Liu et~al.(2024{\natexlab{b}})Liu, Duan, Zhang, Li, Zhang, Zhao, Yuan, Wang, He, Liu, et~al.]{mmbench}
Yuan Liu, Haodong Duan, Yuanhan Zhang, Bo Li, Songyang Zhang, Wangbo Zhao, Yike Yuan, Jiaqi Wang, Conghui He, Ziwei Liu, et~al.
\newblock Mmbench: Is your multi-modal model an all-around player?
\newblock In \emph{European conference on computer vision}, pages 216--233. Springer, 2024{\natexlab{b}}.

\bibitem[Lu et~al.(2023)Lu, Bansal, Xia, Liu, Li, Hajishirzi, Cheng, Chang, Galley, and Gao]{mathvista}
Pan Lu, Hritik Bansal, Tony Xia, Jiacheng Liu, Chunyuan Li, Hannaneh Hajishirzi, Hao Cheng, Kai-Wei Chang, Michel Galley, and Jianfeng Gao.
\newblock Mathvista: Evaluating mathematical reasoning of foundation models in visual contexts.
\newblock \emph{arXiv preprint arXiv:2310.02255}, 2023.

\bibitem[Masry et~al.(2022)Masry, Long, Tan, Joty, and Hoque]{masry2022chartqa}
Ahmed Masry, Do~Xuan Long, Jia~Qing Tan, Shafiq Joty, and Enamul Hoque.
\newblock Chartqa: A benchmark for question answering about charts with visual and logical reasoning.
\newblock \emph{arXiv preprint arXiv:2203.10244}, 2022.

\bibitem[Shao et~al.(2024)Shao, Qian, Xiao, Song, Zong, Wang, Liu, and Li]{visualcot}
Hao Shao, Shengju Qian, Han Xiao, Guanglu Song, Zhuofan Zong, Letian Wang, Yu Liu, and Hongsheng Li.
\newblock Visual cot: Advancing multi-modal language models with a comprehensive dataset and benchmark for chain-of-thought reasoning.
\newblock \emph{Advances in Neural Information Processing Systems}, 37:\penalty0 8612--8642, 2024.

\bibitem[Shen et~al.(2025)Shen, Liu, Li, Fang, Ma, Liao, Shen, Zhang, Zhao, Zhang, et~al.]{vlm-r1}
Haozhan Shen, Peng Liu, Jingcheng Li, Chunxin Fang, Yibo Ma, Jiajia Liao, Qiaoli Shen, Zilun Zhang, Kangjia Zhao, Qianqian Zhang, et~al.
\newblock Vlm-r1: A stable and generalizable r1-style large vision-language model.
\newblock \emph{arXiv preprint arXiv:2504.07615}, 2025.

\bibitem[Team(2025)]{claude-3.5}
Anthropic Team.
\newblock Claude 3.7 sonnet, 2025.

\bibitem[Team et~al.(2024)Team, Georgiev, Lei, Burnell, Bai, Gulati, Tanzer, Vincent, Pan, Wang, et~al.]{gemini-1.5}
Gemini Team, Petko Georgiev, Ving~Ian Lei, Ryan Burnell, Libin Bai, Anmol Gulati, Garrett Tanzer, Damien Vincent, Zhufeng Pan, Shibo Wang, et~al.
\newblock Gemini 1.5: Unlocking multimodal understanding across millions of tokens of context.
\newblock \emph{arXiv preprint arXiv:2403.05530}, 2024.

\bibitem[Team et~al.(2025)Team, Du, Gao, Xing, Jiang, Chen, Li, Xiao, Du, Liao, et~al.]{team2025kimi}
Kimi Team, Angang Du, Bofei Gao, Bowei Xing, Changjiu Jiang, Cheng Chen, Cheng Li, Chenjun Xiao, Chenzhuang Du, Chonghua Liao, et~al.
\newblock Kimi k1.5: Scaling reinforcement learning with llms.
\newblock \emph{arXiv preprint arXiv:2501.12599}, 2025.

\bibitem[Team(2024{\natexlab{a}})]{qvq}
Qwen Team.
\newblock Qvq: To see the world with wisdom, 2024{\natexlab{a}}.

\bibitem[Team(2024{\natexlab{b}})]{qwq}
Qwen Team.
\newblock Qwq: Reflect deeply on the boundaries of the unknown, 2024{\natexlab{b}}.

\bibitem[Wang et~al.(2024)Wang, Chen, Wang, Cao, Liu, Gao, Zhu, Zhu, Lu, Qiao, and Dai]{internvl2.5-mpo}
Weiyun Wang, Zhe Chen, Wenhai Wang, Yue Cao, Yangzhou Liu, Zhangwei Gao, Jinguo Zhu, Xizhou Zhu, Lewei Lu, Yu Qiao, and Jifeng Dai.
\newblock Enhancing the reasoning ability of multimodal large language models via mixed preference optimization.
\newblock \emph{arXiv preprint arXiv:2411.10442}, 2024.

\bibitem[Wang and Peng(2025)]{wang-2025-open-r1-video}
Xiaodong Wang and Peixi Peng.
\newblock Open-r1-video.
\newblock \url{https://github.com/Wang-Xiaodong1899/Open-R1-Video}, 2025.

\bibitem[Wang et~al.(2025{\natexlab{a}})Wang, Wang, Cheng, Fei, Ding, Guo, Tao, and Qiu]{wang2025visuothink}
Yikun Wang, Siyin Wang, Qinyuan Cheng, Zhaoye Fei, Liang Ding, Qipeng Guo, Dacheng Tao, and Xipeng Qiu.
\newblock Visuothink: Empowering lvlm reasoning with multimodal tree search.
\newblock \emph{arXiv preprint arXiv:2504.09130}, 2025{\natexlab{a}}.

\bibitem[Wang et~al.(2025{\natexlab{b}})Wang, Xia, He, Chen, Liu, Zhu, Liang, Wu, Liu, Malladi, et~al.]{charxiv}
Zirui Wang, Mengzhou Xia, Luxi He, Howard Chen, Yitao Liu, Richard Zhu, Kaiqu Liang, Xindi Wu, Haotian Liu, Sadhika Malladi, et~al.
\newblock Charxiv: Charting gaps in realistic chart understanding in multimodal llms.
\newblock \emph{Advances in Neural Information Processing Systems}, 37:\penalty0 113569--113697, 2025{\natexlab{b}}.

\bibitem[Wu et~al.(2025)Wu, Feng, Zhang, Jin, Che, Wen, and Tao]{astar}
Jinyang Wu, Mingkuan Feng, Shuai Zhang, Ruihan Jin, Feihu Che, Zengqi Wen, and Jianhua Tao.
\newblock Boosting multimodal reasoning with mcts-automated structured thinking.
\newblock \emph{arXiv preprint arXiv:2502.02339}, 2025.

\bibitem[Wu et~al.(2024)Wu, Chen, Pan, Liu, Liu, Dai, Gao, Ma, Wu, Wang, et~al.]{deepseek-vl2}
Zhiyu Wu, Xiaokang Chen, Zizheng Pan, Xingchao Liu, Wen Liu, Damai Dai, Huazuo Gao, Yiyang Ma, Chengyue Wu, Bingxuan Wang, et~al.
\newblock Deepseek-vl2: Mixture-of-experts vision-language models for advanced multimodal understanding.
\newblock \emph{arXiv preprint arXiv:2412.10302}, 2024.

\bibitem[Xie et~al.(2025)Xie, Gao, Ren, Luo, Hong, Dai, Zhou, Qiu, Wu, and Luo]{logic-rl}
Tian Xie, Zitian Gao, Qingnan Ren, Haoming Luo, Yuqian Hong, Bryan Dai, Joey Zhou, Kai Qiu, Zhirong Wu, and Chong Luo.
\newblock Logic-rl: Unleashing llm reasoning with rule-based reinforcement learning.
\newblock \emph{arXiv preprint arXiv:2502.14768}, 2025.

\bibitem[Xu et~al.(2024)Xu, Jin, Hao, Song, Sun, and Yuan]{llava-cot}
Guowei Xu, Peng Jin, Li Hao, Yibing Song, Lichao Sun, and Li Yuan.
\newblock Llava-o1: Let vision language models reason step-by-step.
\newblock \emph{arXiv preprint arXiv:2411.10440}, 2024.

\bibitem[Yao et~al.(2024{\natexlab{a}})Yao, Huang, Wu, Zhang, Wang, Liu, Wang, Song, Feng, Shen, et~al.]{mulberry}
Huanjin Yao, Jiaxing Huang, Wenhao Wu, Jingyi Zhang, Yibo Wang, Shunyu Liu, Yingjie Wang, Yuxin Song, Haocheng Feng, Li Shen, et~al.
\newblock Mulberry: Empowering mllm with o1-like reasoning and reflection via collective monte carlo tree search.
\newblock \emph{arXiv preprint arXiv:2412.18319}, 2024{\natexlab{a}}.

\bibitem[Yao et~al.(2025)Yao, Yin, Zhang, Yang, Wang, Wu, Su, Shen, Qiu, Tao, et~al.]{R1-ShareVL}
Huanjin Yao, Qixiang Yin, Jingyi Zhang, Min Yang, Yibo Wang, Wenhao Wu, Fei Su, Li Shen, Minghui Qiu, Dacheng Tao, et~al.
\newblock R1-sharevl: Incentivizing reasoning capability of multimodal large language models via share-grpo.
\newblock \emph{arXiv preprint arXiv:2505.16673}, 2025.

\bibitem[Yao et~al.(2024{\natexlab{b}})Yao, Yu, Zhang, Wang, Cui, Zhu, Cai, Li, Zhao, He, et~al.]{minicpm-v}
Yuan Yao, Tianyu Yu, Ao Zhang, Chongyi Wang, Junbo Cui, Hongji Zhu, Tianchi Cai, Haoyu Li, Weilin Zhao, Zhihui He, et~al.
\newblock Minicpm-v: A gpt-4v level mllm on your phone.
\newblock \emph{arXiv preprint arXiv:2408.01800}, 2024{\natexlab{b}}.

\bibitem[Ying et~al.(2024)Ying, Meng, Wang, Li, Lin, Yang, Zhang, Zhang, Lin, Liu, et~al.]{mmtbench}
Kaining Ying, Fanqing Meng, Jin Wang, Zhiqian Li, Han Lin, Yue Yang, Hao Zhang, Wenbo Zhang, Yuqi Lin, Shuo Liu, et~al.
\newblock Mmt-bench: A comprehensive multimodal benchmark for evaluating large vision-language models towards multitask agi.
\newblock \emph{arXiv preprint arXiv:2404.16006}, 2024.

\bibitem[Yu et~al.(2023)Yu, Yang, Li, Wang, Lin, Liu, Wang, and Wang]{mmvet}
Weihao Yu, Zhengyuan Yang, Linjie Li, Jianfeng Wang, Kevin Lin, Zicheng Liu, Xinchao Wang, and Lijuan Wang.
\newblock Mm-vet: Evaluating large multimodal models for integrated capabilities.
\newblock \emph{arXiv preprint arXiv:2308.02490}, 2023.

\bibitem[Yue et~al.(2024{\natexlab{a}})Yue, Ni, Zhang, Zheng, Liu, Zhang, Stevens, Jiang, Ren, Sun, et~al.]{mmmu}
Xiang Yue, Yuansheng Ni, Kai Zhang, Tianyu Zheng, Ruoqi Liu, Ge Zhang, Samuel Stevens, Dongfu Jiang, Weiming Ren, Yuxuan Sun, et~al.
\newblock Mmmu: A massive multi-discipline multimodal understanding and reasoning benchmark for expert agi.
\newblock In \emph{Proceedings of the IEEE/CVF Conference on Computer Vision and Pattern Recognition}, pages 9556--9567, 2024{\natexlab{a}}.

\bibitem[Yue et~al.(2024{\natexlab{b}})Yue, Zheng, Ni, Wang, Zhang, Tong, Sun, Yu, Zhang, Sun, et~al.]{mmmu_pro}
Xiang Yue, Tianyu Zheng, Yuansheng Ni, Yubo Wang, Kai Zhang, Shengbang Tong, Yuxuan Sun, Botao Yu, Ge Zhang, Huan Sun, et~al.
\newblock Mmmu-pro: A more robust multi-discipline multimodal understanding benchmark.
\newblock \emph{arXiv preprint arXiv:2409.02813}, 2024{\natexlab{b}}.

\bibitem[Zhang et~al.(2024{\natexlab{a}})Zhang, Gao, Gan, Dufter, Wenzel, Huang, Shah, Du, Zhang, Li, et~al.]{zhang2024mm1}
Haotian Zhang, Mingfei Gao, Zhe Gan, Philipp Dufter, Nina Wenzel, Forrest Huang, Dhruti Shah, Xianzhi Du, Bowen Zhang, Yanghao Li, et~al.
\newblock Mm1. 5: Methods, analysis \& insights from multimodal llm fine-tuning.
\newblock \emph{arXiv preprint arXiv:2409.20566}, 2024{\natexlab{a}}.

\bibitem[Zhang et~al.(2025)Zhang, Huang, Yao, Liu, Zhang, Lu, and Tao]{r1-vl}
Jingyi Zhang, Jiaxing Huang, Huanjin Yao, Shunyu Liu, Xikun Zhang, Shijian Lu, and Dacheng Tao.
\newblock R1-vl: Learning to reason with multimodal large language models via step-wise group relative policy optimization.
\newblock \emph{arXiv preprint arXiv:2503.12937}, 2025.

\bibitem[Zhang et~al.(2024{\natexlab{b}})Zhang, Jiang, Zhang, Lin, Guo, Qiu, Zhou, Lu, Chang, Qiao, et~al.]{mathverse}
Renrui Zhang, Dongzhi Jiang, Yichi Zhang, Haokun Lin, Ziyu Guo, Pengshuo Qiu, Aojun Zhou, Pan Lu, Kai-Wei Chang, Yu Qiao, et~al.
\newblock Mathverse: Does your multi-modal llm truly see the diagrams in visual math problems?
\newblock In \emph{European Conference on Computer Vision}, pages 169--186. Springer, 2024{\natexlab{b}}.

\bibitem[Zhang et~al.(2024{\natexlab{c}})Zhang, Zhang, Li, Zhang, Sun, Gan, Yang, Pang, and Yang]{llava-reasoner}
Ruohong Zhang, Bowen Zhang, Yanghao Li, Haotian Zhang, Zhiqing Sun, Zhe Gan, Yinfei Yang, Ruoming Pang, and Yiming Yang.
\newblock Improve vision language model chain-of-thought reasoning.
\newblock \emph{arXiv preprint arXiv:2410.16198}, 2024{\natexlab{c}}.

\bibitem[Zou et~al.(2024)Zou, Guo, Yang, Zhang, Hu, and Zhang]{dynamath}
Chengke Zou, Xingang Guo, Rui Yang, Junyu Zhang, Bin Hu, and Huan Zhang.
\newblock Dynamath: A dynamic visual benchmark for evaluating mathematical reasoning robustness of vision language models.
\newblock \emph{arXiv preprint arXiv:2411.00836}, 2024.

\end{thebibliography}
}

\end{document}